\documentclass[conference]{IEEEtran}

\makeatletter
\def\ps@IEEEtitlepagestyle{%
  \def\@oddfoot{\mycopyrightnotice}%
  \def\@oddhead{\hbox{}\@IEEEheaderstyle\leftmark\hfil\thepage}\relax
  \def\@evenhead{\@IEEEheaderstyle\thepage\hfil\leftmark\hbox{}}\relax
  \def\@evenfoot{}%
}
\def\mycopyrightnotice{%
  \begin{minipage}{\textwidth}
  \scriptsize
  \copyright~2021 IEEE. Personal use of this material is permitted. Permission from IEEE must be obtained for all other uses, in any current or future media, including reprinting/republishing this material for advertising or promotional purposes, creating new collective works, for resale or redistribution to servers or lists, or reuse of any copyrighted component of this work in other works. 
  
  This work has been accepted at the 2021 32nd IEEE International Conference on Application-specific Systems, Architectures and Processors (ASAP).
  \end{minipage}
}
\makeatother

\usepackage{booktabs} 
\usepackage{enumerate}
\usepackage{graphicx}
\usepackage{amsmath}
\usepackage{multirow}
\usepackage{footnote}
\usepackage{threeparttable}
\usepackage{amsmath}
\usepackage{textcomp}
\usepackage{lipsum}
\usepackage{hyperref}
\usepackage{nicefrac}

\newcommand{\figref}[1]{Fig.~\ref{#1}}
\newcommand{\eqnref}[1]{Equation~(\ref{#1})}
\newcommand{\secref}[1]{Section~\ref{#1}}

\newcommand{\tabref}[1]{Table~\ref{#1}}

\usepackage[square, comma, numbers]{natbib}
\usepackage[ruled,linesnumbered]{algorithm2e}
\usepackage{algpseudocode}

\usepackage{graphicx}
\usepackage{subcaption}
\usepackage[dvipsnames]{xcolor}

\begin{document}
%

\title{Accelerating Recurrent Neural Networks for Gravitational Wave Experiments}

 \author{
\IEEEauthorblockN{
Zhiqiang Que\IEEEauthorrefmark{1}, 
Erwei Wang\IEEEauthorrefmark{1}, 
Umar Marikar\IEEEauthorrefmark{1}, 
Eric Moreno\IEEEauthorrefmark{2},
Jennifer Ngadiuba\IEEEauthorrefmark{2}, \\
Hamza Javed\IEEEauthorrefmark{3},
Bartłomiej Borzyszkowski\IEEEauthorrefmark{3},
Thea Aarrestad\IEEEauthorrefmark{3},
Vladimir Loncar\IEEEauthorrefmark{3},
Sioni Summers\IEEEauthorrefmark{3},\\
Maurizio Pierini\IEEEauthorrefmark{3},
Peter Y Cheung\IEEEauthorrefmark{1},
Wayne Luk\IEEEauthorrefmark{1}
}

\IEEEauthorblockA{
\IEEEauthorrefmark{2}
California Institute of Technology, Pasadena, CA, USA}

\IEEEauthorblockA{
\IEEEauthorrefmark{3}
European Organization for Nuclear
Research (CERN), Geneva, Switzerland, \\
\{jennifer.ngadiuba, sioni.paris.summers, Maurizio.Pierini\}@cern.ch
}

\IEEEauthorblockA{
\IEEEauthorrefmark{1}Imperial College London, UK, 
\{z.que, w.luk\}@imperial.ac.uk}

 {\vspace{-0.5cm}}
}

\maketitle



\begin{abstract}
This paper presents novel reconfigurable architectures for reducing the latency of recurrent neural networks (RNNs) that are used for detecting gravitational waves. Gravitational interferometers such as the LIGO detectors capture cosmic events such as black hole mergers which happen at unknown times and of varying durations, producing time-series data.
We have developed a new architecture capable of accelerating RNN inference for analyzing time-series data from LIGO detectors. This architecture is based on optimizing the initiation intervals (II) in a multi-layer LSTM (Long Short-Term Memory) network, by identifying appropriate reuse factors for each layer. A customizable template for this architecture has been designed, which enables the generation of low-latency FPGA designs with efficient resource utilization using high-level synthesis tools. The proposed approach has been evaluated based on two LSTM models, targeting a ZYNQ 7045 FPGA and a U250 FPGA. Experimental results show that with balanced II, the number of DSPs can be reduced up to 42\% while achieving the same IIs. When compared to other FPGA-based LSTM designs, our design can achieve about 4.92 to 12.4 times lower latency. 
\end{abstract}


\section{Introduction}
Recurrent Neural Networks (RNNs) are a type of architecture specialized for processing ordered data, for example time-series data. These networks 
have applications in speech recognition~\cite{han2017ese}, DNA sequence analysis, and physics experiments~\cite{schmitt2019investigating, lin2021detection}. 
An exciting physics experiment concerns the detection of gravitational waves, predicted by Albert Einstein a hundred years ago. The first detected wave came from a collision between two black holes, reaching the earth after 1.3 billion years. 
The detectors at the Laser Interferometer Gravitational-Wave Observatory (LIGO) produce time-series data, as they capture cosmic events such as black hole mergers which happen at unknown times and of varying durations.
Accelerating RNN inference using reconfigurable accelerators such as FPGAs would enable sophisticated processing, such as anomaly detection, to run in real time on the data stream from the detector and generate a fast response. 
Among the many RNN variants, the most popular one is Long Short-Term Memory (LSTM). FPGAs have been used to speed up the inference of RNNs/LSTMs \cite{han2017ese, guan2017fpga, fowers2018configurable, eriko_fccm19, que2020optimizing}, which offer benefits of low latency and low power consumption compared to CPUs or GPUs. 

However, existing LSTM accelerators cannot support low-latency and effective multi-layer execution, especially when targeting small LSTM models with requirements of ultra low latency and ultra high throughput for scientific applications. Many existing FPGA-based LSTM accelerators are designed with the same idea as their GPU counterparts, which utilize a single computational engine architecture where the engine is designed to run one block or layer at one time, and the whole network is processed by running the engine repeatedly~\cite{fowers2018configurable,eriko_fccm19}. 
Their design consists of arranging computing resources to form a single core with many processing elements,
leveraging data level parallelism. For example, Brainwave~\cite{fowers2018configurable} is a single-threaded neural processing unit (NPU) which has 96,000 processing elements (PEs). However, when the size of the targeted LSTM layer is small, these hardware resources will not be fully utilized, e.g., when targeting a small LSTM layer, the Brainwave hardware utilization is lower than 1\%~\cite{fowers2018configurable}, while the utilization of the NPU can be lower than 15\%~\cite{eriko_fccm19}.
Moreover, since a single engine is used, 
the various layers must have the same amount of parallelism which is not flexible to take full advantage of the customizability of FPGAs.
Thus, this work applies a layer-wise architecture to map all the LSTM layers on-chip and perform the computation for different layers on their own unit with independent optimization to achieve low latency and high system throughput. 


\begin{figure}
\begin{center}
\includegraphics[width=0.9\linewidth]{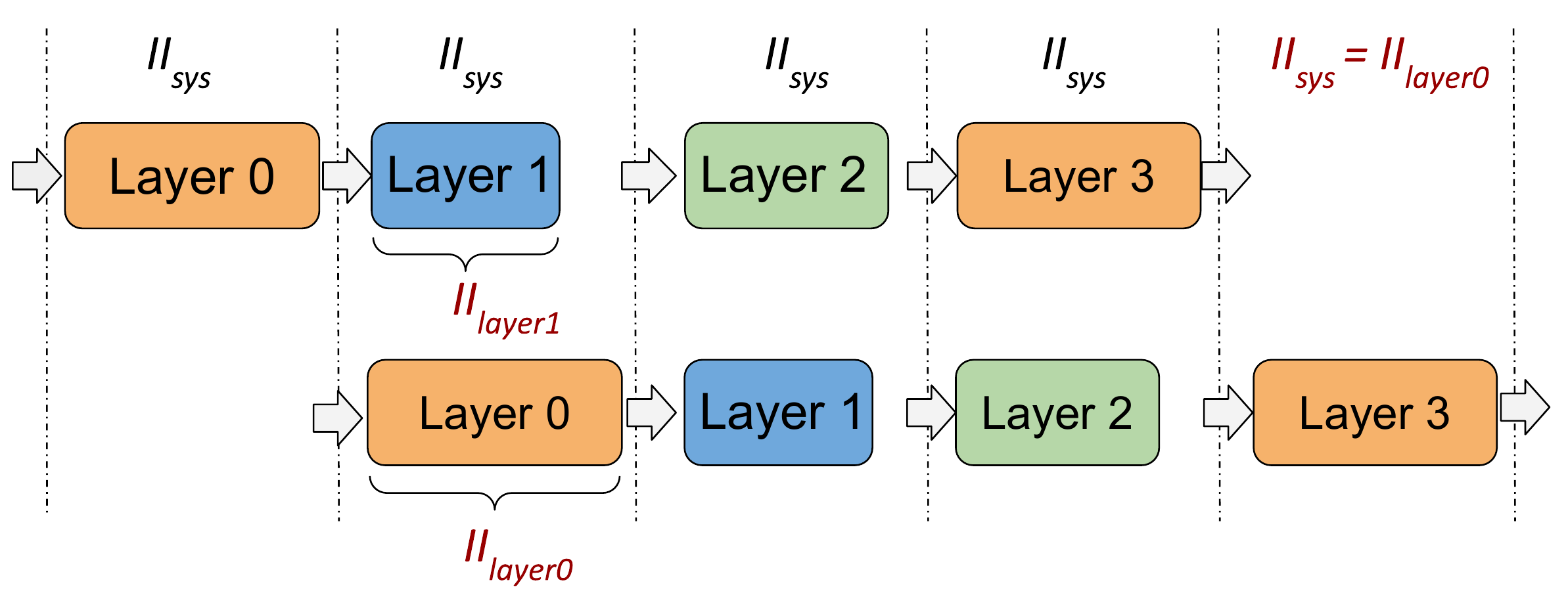}
\end{center}
   \caption{Unbalanced layer IIs among various cascaded layers in an RNN model}
\label{fig:layer_pipeline}
\end{figure}

Unlike CNN inference designs~\cite{zhang2018dnnbuilder, nakahara2020high} which only have forward datapaths and can be fully pipelined, there are feedback datapaths in RNN inference and data dependencies exist between the current timestep and the next timestep. 
Unrolling the timesteps fully may help, however the sequence length (timestep) of an LSTM model is usually larger than the number of layers~\cite{peng2019exploiting}, e.g., 1500 timesteps in an LSTM layer in DeepSpeech~\cite{deepspeech}, which makes the full unrolling of timesteps impractical on FPGAs because of the limited hardware resources.


To accelerate an RNN model with multiple LSTM layers,
this work proposes coarse grained pipelining with balanced II
(initiation interval) to improve system throughput and reduce
latency. This is achieved by identifying appropriate reuse factors for each layer, resulting in fast response and enhanced resolution for processing sensor data.
It can achieve the best (smallest) system level II for a neural network with multiple LSTM layers on a given FPGA. The II is the number of clock cycles before a unit can accept new inputs and is generally the most critical performance metric in systems~\cite{xilinx_sdsoc}. A perfect pipeline has $II=1$ cycle, as this is required to keep all pipeline stages busy. However, the II of an LSTM layer is generally larger than one because of the data dependencies.
For a model with multiple layers in sequence, the initiation interval of this model is decided by the largest II among all the layers~\cite{de2020transformations}, as shown in~\figref{fig:layer_pipeline}. The unbalanced IIs in various layers result in hardware inefficiency and low throughput. 
Accelerating a deep LSTM model is challenging since the computation load varies greatly among layers and data dependency exists both time-wise and layer-wise. 


Our approach is to ensure all the layer IIs are balanced to
eliminate system stall, so that the system becomes a coarse
grained seamless pipeline. It increases pipeline parallelism
by performing more computations without increasing latency,
and without introducing additional memory traffic or
storage.
Unbalanced IIs in a pipeline is a common issue,  but few studies address balancing IIs in the context of accelerating multi-layer DNNs, especially for RNNs/LSTMs. The proposed coarse-grained pipelining is similar to layer parallelism but the granularity in our approach does not need to cover an entire layer. An LSTM layer can still be divided into multiple blocks with pipeline parallelism. In addition, a customizable template for this architecture has been designed, which enables the generation of low-latency FPGA designs with efficient resource utilization using high-level synthesis (HLS) tools. Moreover, We develop an optimization algorithm such that, given the dimensions of the LSTM layers and a resource budget, computes a partitioning of the FPGA resources for an efficient

To the best of our knowledge, this is the first work to propose balancing IIs for a coarse-grained pipelined architecture to enable fast multi-layer LSTM data analysis in gravitational wave experiments. This work could help improve performance of next generation Gravitational Wave detectors.


We make the following contributions in this paper:
\begin{itemize}
\item A novel technique for balancing IIs of multi-layer LSTM
inference to increase hardware efficiency and system
throughput for data analysis in gravitational wave
experiments.
\item A scalable and low latency LSTM template which enables the generation of low-latency FPGA designs with efficient resource utilization by HLS tools. We open source the templates with some examples\footnote{https://github.com/walkieq/RNN\_HLS}. 
\item A comprehensive evaluation of the proposed method and hardware architecture.
\end{itemize}

The specific RNN layered structure and coefficients are LIGO specific, but the need for low latency would benefit many other applications, especially those requiring real-time response, e.g., low latency would benefit the Large Hadron Collider (LHC) physics~\cite{duarte2018fast}, adaptive radiotherapy~\cite{thorwarth2021technical} and electronic trading~\cite{denholm2014low}. The proposed techniques can be adapted to address these other applications. 



\section{Background and Preliminaries}


\begin{figure}
\begin{center}
\includegraphics[width=0.9\linewidth]{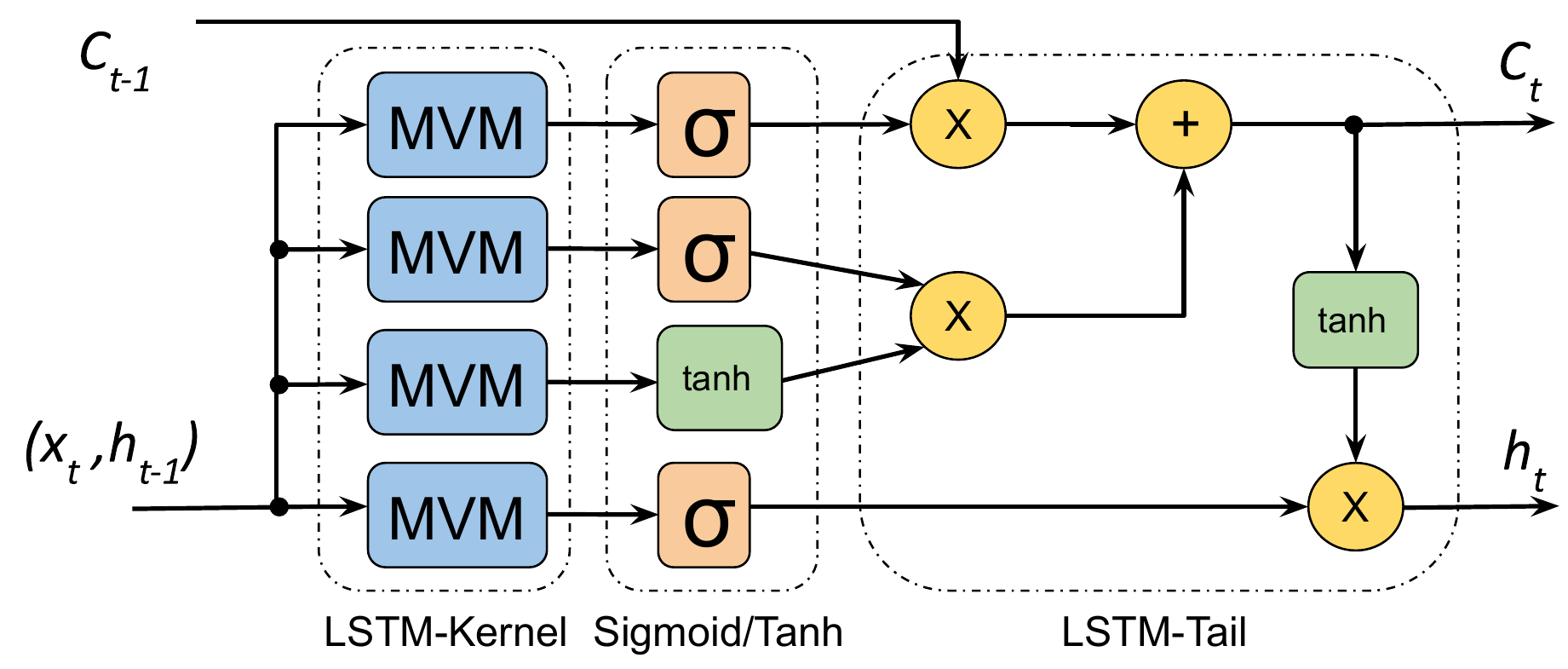}
\end{center}
   \caption{Structure of an LSTM cell}
\label{fig:lstm}
\end{figure}
RNNs/LSTMs have been shown to have useful properties with many significant applications. 
This study follows the standard LSTM cell \cite{guan2017fpga, fowers2018configurable, eriko_fccm19}. 
\ref{fig:lstm} shows an LSTM cell.
It consists of three main parts. At the front, there are four LSTM gates which perform matrix-vector multiplication (MVM), followed by activation functions. While in the tail, there are a few element-wise operations. 
The hidden state $h_t$, which will be fed back from the tail to the front, is produced by the following equations:
\begin{align*}
    i_t &=\sigma(W_i[x_t, h_{t-1}] + b_i), 
    & f_t &=\sigma(W_f[x_t, h_{t-1}] + b_f) \\
    g_t &= \text{tanh}(W_g[x_t, h_{t-1}] + b_u), 
    & o_t &= \sigma(W_o[x_t, h_{t-1}] + b_o) \\
    c_t &= f_t \odot c_{t-1} + i_t \odot g_t, 
    & h_t &= o_t \odot tanh(c_t)
\end{align*}
Here, $\sigma$, $tanh$ and $\odot$ stand for the sigmoid function, the hyperbolic tangent function and element-wise multiplication respectively. $i, f, g$ and $o$ represent the input, forget, input modulation and output gate respectively. The input modulation gate is often considered as a sub-part of the input gate. The input vector and hidden vector are combined so that $W$ represents the weight matrix for both vectors. Bias term is represented as $b$. The output $c_t$ is the internal memory cell state and $h_t$ is the output of the cell, also called the hidden vector, which is passed to the next timestep or next layer. 


\section{Design and Optimization Methodology} \label{sec:design}
\begin{table}
\centering
\begin{threeparttable}
\caption{System Parameters}
\label{table:parameters}
\renewcommand{\arraystretch}{1.2}

\begin{tabular}{|l | l|}

\hline 
$II_{sys}$ & System initiation interval \\
\hline
$TS$  & Timestep number\\
\hline 
$ii_{N}$ & Timestep loop initiation interval in the LSTM layer $N$ \\
\hline 
$II_{N}$ & Initiation interval for layer $N$ \\
\hline
$LT_{N}$  & Latency of a single timestep loop for layer $N$ \\
\hline
$LT_{\alpha}$  & Latency of the unit $\alpha$; $\alpha$ could be mult / mvm / tail / $\sigma$ \\
\hline
$x_t$ & The input vector $x$ at timestep $t$\\
\hline 
$h_t$ & The hidden vector $h$ at timestep $t$ \\
\hline 
${Wx}$ & LSTM gates weight matrix for input vector.\\
\hline 
${Wh}$ & LSTM gates weight matrix for hidden vector.\\
\hline
$Lx$ & Number of elements in the input vector $x$\\
\hline
$Lh$ & Number of elements in the hidden vector $h$\\
\hline 
$R_x$  & Reuse factor for MVM involving LSTM input vector $x_t$\\
\hline 
$R_h$  & Reuse factor for MVM involving LSTM hidden vector $h_t$\\
\hline 
$R_t$  & Reuse factor for LSTM tail unit\\
\hline 

\end{tabular}
\end{threeparttable}
\end{table}

This section analyzes unbalanced II issues and introduces several optimizations for multi-layer RNN designs. 
We define a few parameters, as shown in~\tabref{table:parameters} for later calculations.

\subsection{LSTM-based autoencoder for gravitational wave detection} \label{sec:autoencoder}
\begin{figure}
\begin{center}
\includegraphics[width=0.8\linewidth]{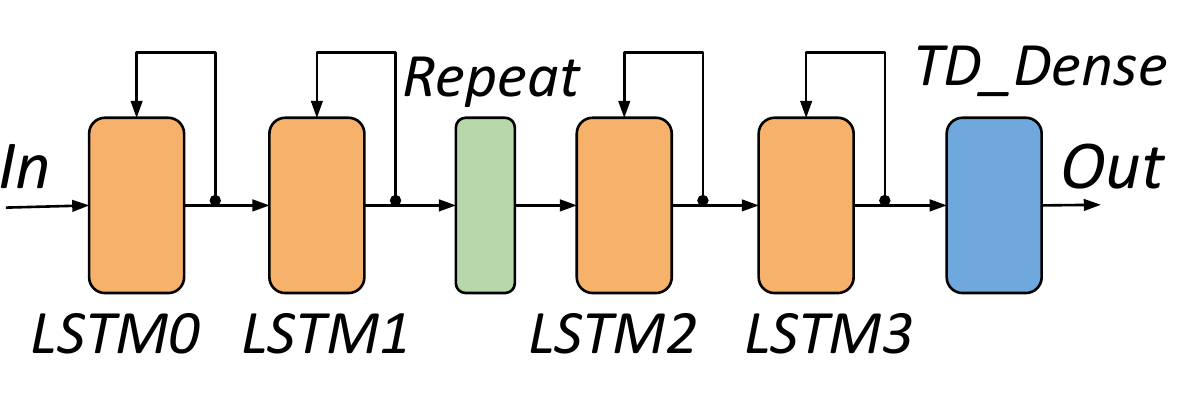}
\end{center}
   \caption{Overview of the LSTM-based autoencoder}
\label{fig:ae}
\end{figure}

\figref{fig:ae} shows an overview of the LSTM-based autoencoder used for gravitational wave detection. The models and the dataset are available on GitHub~\cite{ligo_anomaly,ligo_paper}.
The autoencoder consists of two components, an encoder and decoder. The encoder learns to transform data from the input layer into a latent-space representation, which acts as a data "bottleneck". The decoder then reconstructs the output of the reduced latent representation as close as possible to its original input. When the error between input and reconstructed values is high, the input is flagged as anomalous. In this work, an LSTM-based autoencoder is used as an unsupervised prediction model to detect the anomalies for gravitational waves. This works by only training the LSTM-autoencoder to encode and decode normal background conditions at the LIGO interferometers. When an event containing a gravitational wave passes through the autoencoder, the model cannot encode and decode the additional strain provided by the gravitational wave. Both the encoder and decoder have two LSTM layers. A TimeDistributed dense layer is applied before the data output. 

\subsection{System II for multi-layer LSTM networks }\label{subsec:II}

Accelerating a deep LSTM model which has multiple layers is challenging since the computation varies greatly among layers and data dependencies exist both time-wise and layer-wise. 
An efficient technique to improve throughput and reuse computational resources is to pipeline hardware units. If each input can overlap with itself, we can achieve simultaneously inference parallelism within a run by coarse grained pipelining as shown in~\figref{fig:layer_pipeline}.

\begin{figure}
\begin{center}
\includegraphics[width=0.9\linewidth]{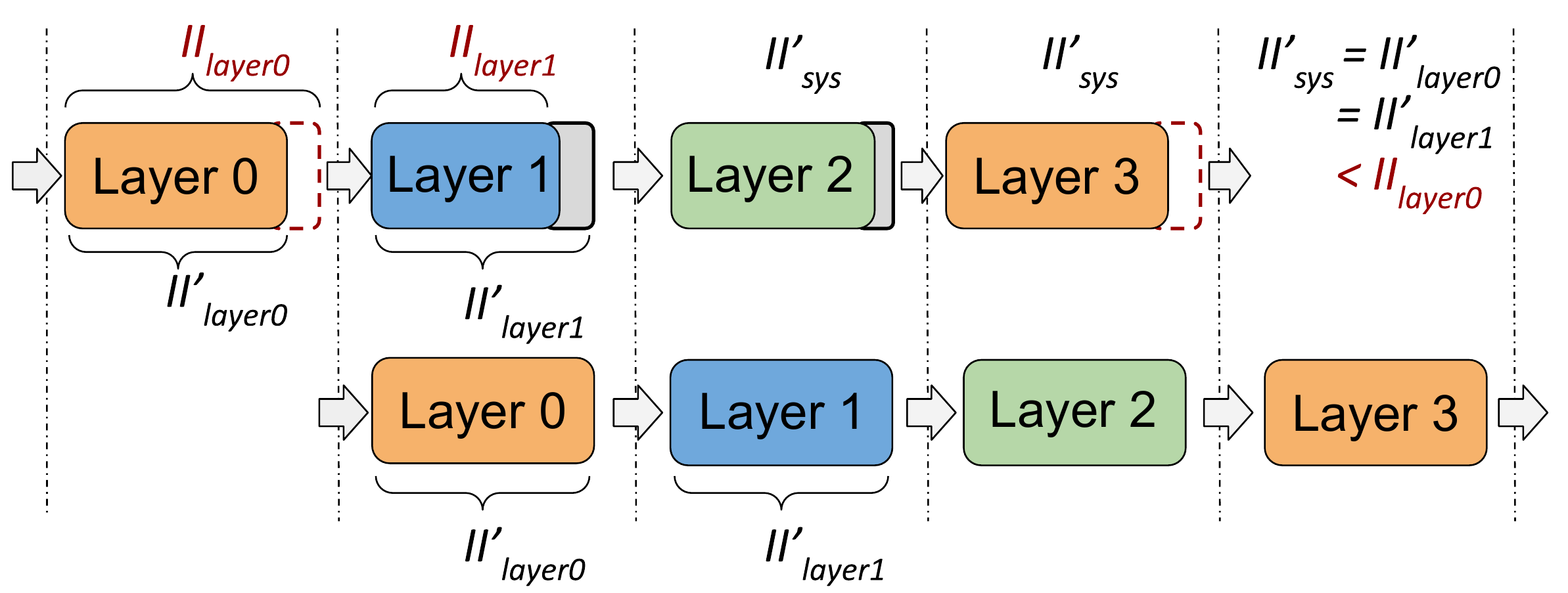}
\end{center}
   \caption{Overview of the method used to balance IIs}
\label{fig:II_balancing}
\end{figure}

However, a naive implementation can result in a large number of idle cycles due to inter-layer dependencies since the pipeline is not seamless; a particular layer might stall until the previous layer finishes. The unbalanced IIs in various layers results in hardware inefficiency and low system throughput. Typically, the particular layer with the largest II should be optimized since it dominates the system II. Generally, the II cycles can be reduced if more hardware resources are allocated to that particular layer by adding more parallelisms. So the targeted layer should be allocated as many hardware resources as possible. However, the hardware resources on a given FPGA is limited, which means that the other layers may occupy less hardware resources. When the resources for a layer decrease, the II of that layer will increase. Then this layer may become the one that has the largest II and dominates the design. Thus, the optimal case is that all the layers have the same II, in which scenario the design utilizes the hardware resources efficiently and achieves the highest system throughput as shown in~\figref{fig:II_balancing}. 

Besides, we find that we do not need to unroll every unit in order to achieve the lowest II. Some hardware resources can be saved from the units which do not require full unrolling. And then these saved hardware resources can be reallocated to the other units which dominate the system to achieve low initiation intervals. As shown in~\figref{fig:II_balancing}, the hardware resources for layer~1 can be reduced so that the saved resources can be reallocated for layer~0. The $II_{layer1}$ is increased to $II'_{layer1}$ while the $II_{layer0}$ which is the largest can be reduced to $II'_{layer0}$ so that the final system $II_{sys}$ can be reduced.

Partitioning FPGA resources to enhance throughput has been studied for CNNs~\cite{zhang2018dnnbuilder, nakahara2020high, shen2017maximizing, zhang2020dnnexplorer} but they do not touch the RNNs and the recurrent nature as well as the data dependencies in RNN computations, which are absent from CNNs. 
We develop an optimization algorithm such that, given the dimensions of the LSTM layers and a resource budget, computes a partitioning of the FPGA resources for an efficient
and balanced high-performance design. Our algorithm runs in seconds and produces a set of reuse factors~\cite{duarte2018fast}. We then use these factors to parameterize an LSTM template design specified using HLS to form a complete multi-layer LSTM implementation.
Since all the layers have the same II, we only need to focus on the optimization for a single LSTM layer. The layer II and system II are
\begin{align*}
    II_N &= ii_N \times TS  \tag{1} \label{eq:1} \\
    II_{sys} &= \text{max}(II_0,\ II_1,\ ...,\ II_N ) \tag{2} \label{eq:2}
\end{align*}
The original $II_N$ should be $II_N = ii_N \times TS + (LT_N-ii_N)$. However, the extra $(LT_N-ii_N)$ cycles can be eliminated after using the rewind for Vivado\_HLS \textit{\#pragma pipeline}. The rewind is an optional keyword that enables rewinding, or continuous loop pipelining with no pause between the end of one loop iteration and the start of the next iteration. 
So the proposed balancing method has two benefits. First, it improves throughput due to pipelining. Second, it reduces system latency since if the LSTM loop initiation interval, $ii_N$, can be reduced by 1 cycle, then the system latency can be reduced by $TS$ cycles in total according to~\eqnref{eq:1}.



\subsection{The II of a single LSTM layer}

\begin{figure}
\begin{center}
\includegraphics[width=0.8\linewidth]{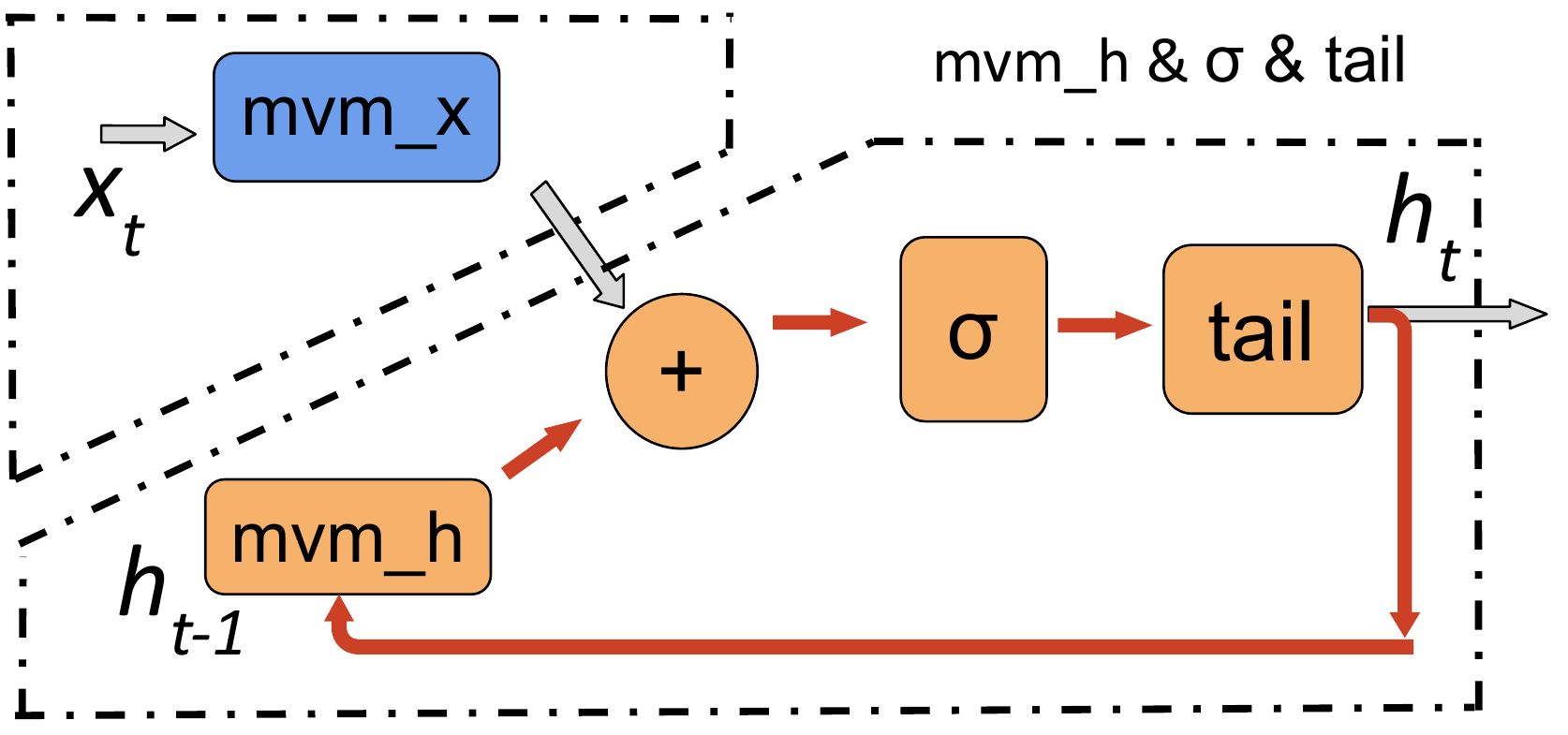}
\end{center}
   \caption{An LSTM layer after performing the transformation}
\label{fig:lstm_tran}
\end{figure}

This work splits one LSTM layer into two sub-layers. The first one is the $mvm\_x$ which has no data dependencies and performs MVM operations for the LSTM gates involving the input vectors while the second one includes all the others which form a loop with data dependencies, as shown in~\figref{fig:lstm_tran}. 
For accelerating LSTM layers used for gravitational wave detection, the system is designed to achieve the average latency (system II) as small as possible. To achieve the lowest system II, fully unrolling the neural network model is an effective method which utilizes a multiplier only once in the computation of a layer. E.g., a fully connected (FC) layer with input size $num\_in$ and output size $num\_out$ can achieve the lowest latency if there are $num\_in \times num\_out$ multipliers. This is the most parallel and fast way a layer can be computed. It has been demonstrated in the HLS4ML based DNN designs for particle physics~\cite{duarte2018fast}. However, unlike forward computation in the FC layers used in the design of~\cite{duarte2018fast}, there are data dependencies in LSTM computations. 

\begin{figure}
\begin{center}
\includegraphics[width=0.9\linewidth]{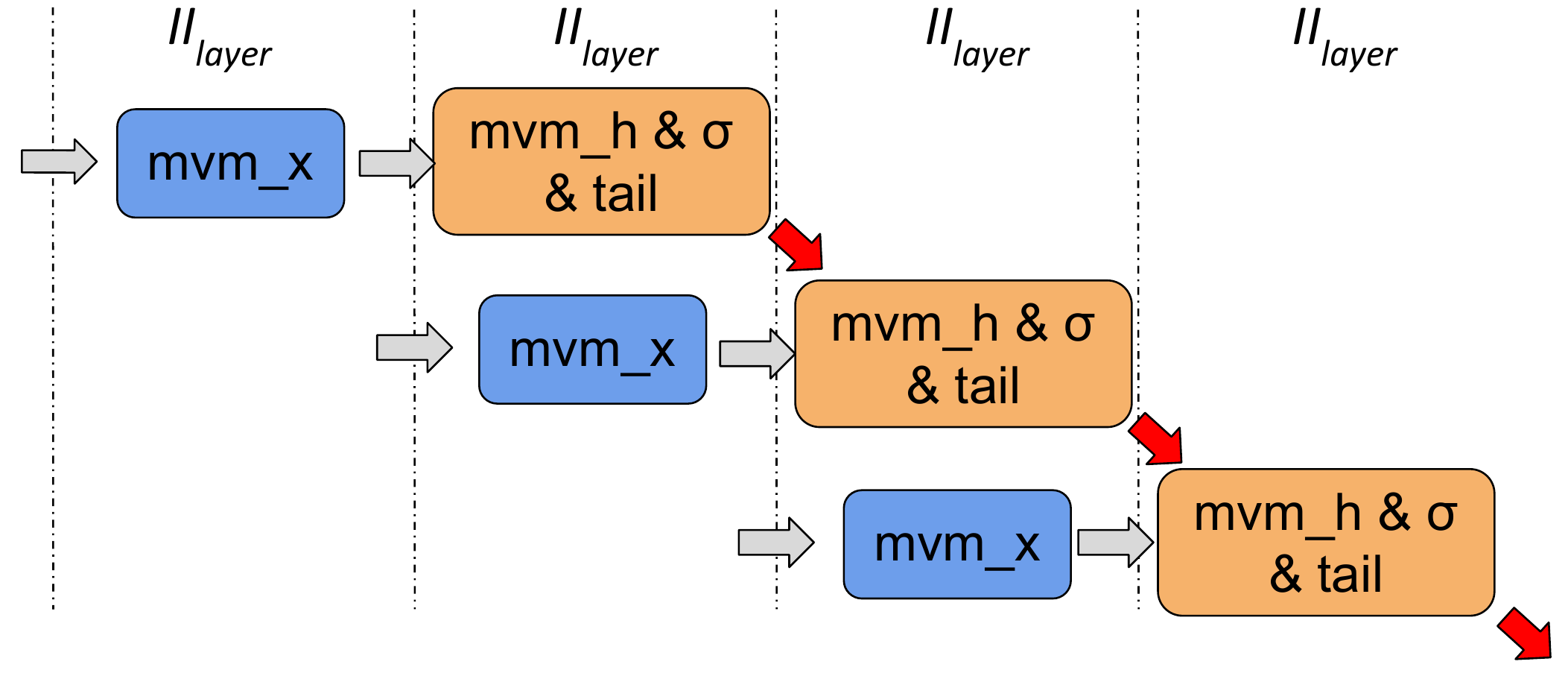}
\end{center}
   \caption{Coarse grained pipelining in an LSTM layer}
\label{fig:ts_pipeline}
\end{figure}

After we have split the LSTM layer into two sub-layers, the two can be pipelined as shown in~\figref{fig:ts_pipeline}. According to the discussion in~\secref{subsec:II}, the optimal case is when the two sub-layers have the same II. Since the second sub-layer is complex and its II is usually larger than the one of the first sub-layer, the parallelism for the first sub-layer does not need to be as large as possible, resulting in a reduction of the number of multipliers needed to process the $mvm\_x$ unit. The saved multipliers can then be reallocated for other layers to achieve a lower system II. Reducing the parallelism of $mvm\_x$ does not hurt the system latency. Normally, each input vector can finish the calculation in the shadow region of processing the $h_t$ because of the pipelining. Besides, the cycles for processing the first $mvm\_x$ can be eliminated when calculating the layer II because of the keyword of rewind in Vivado HLS. 


While the second sub-layer may seem complex, if the design is split into more sub-layers, these sub-layers cannot be coarse grained pipelined. The reason is that the start of the next iteration needs the result from the current iteration, as shown by the red arrows in~\figref{fig:ts_pipeline}

\begin{figure}
\begin{center}
\includegraphics[width=0.9\linewidth]{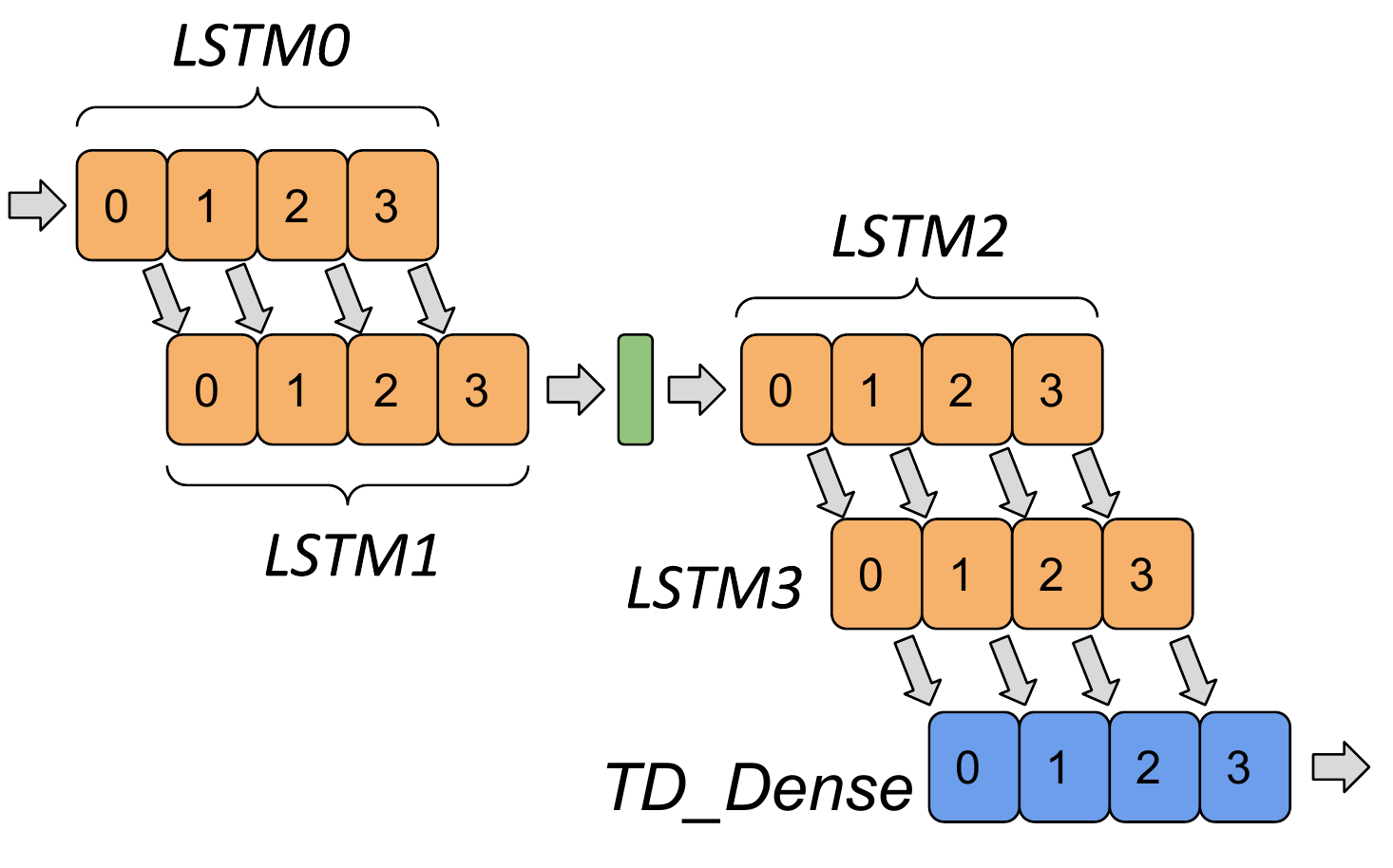}
\end{center}
   \caption{Timestep overlapping}
\label{fig:ae_ts_pipeline}
\end{figure}

\subsection{Overlapping the computations in cascaded LSTM layers}
In the proposed coarse grained pipelining, the processing of the cascaded LSTM layers can be overlapped. The second layer does not need to wait for the whole sequence of hidden vectors to be ready. Just one hidden vector from the former LSTM layer is sufficient to start the calculation of the next LSTM layer as shown in~\figref{fig:ae_ts_pipeline}. It helps to reduce the overall system latency. 
It has to be noted that the LSTM2 can only start after the LSTM1 calculation is completed, since only the last timestep hidden vector is returned in LSTM1, which is decided by the structure of the autoencoder.  



\section{Implementation}

\subsection{HLS implementation}\label{subsec:HLS_implementation}


This work maps all the layers on-chip and different layers run in a fashion of coarse grained pipelining to increase the system throughput. Besides, this work always seeks to achieve extremely low latency by utilizing as many hardware resources as possible. 
However, because of the data dependencies between different timesteps in LSTM calculation, the initiation interval is typically larger than 1. In this case, HLS will automatically increase the initiation interval until it can find a feasible schedule. 
For complex codes it is common to partition functionality into multiple modules, streaming data between them through explicit interfaces. Smaller components are more modular, making them easier to reuse, debug and verify. The effort required by the HLS tool to schedule code sections increases dramatically with a large number of operations that need to be considered for the dependency and pipelining analysis. Scheduling logic in smaller chunks is thus beneficial for compilation time and sometimes also for system latency. 
Our experiments show that inlining every function, especially the $mvm\_x$ and $mvm\_h$ in the LSTM gates , brings large II when the involved matrices are large. 


The trade-off between latency, throughput and FPGA resource usage is determined by the parallelization of the inference calculation. This work adopts the reuse factor used in~\cite{duarte2018fast} to fine tune the parallelism, which is configured to set the number of times a multiplier is used in the computation of a module. In one extreme, all multiplications can be performed simultaneously using a maximal number of multipliers, while alternatively in the other extreme, one can use only one multiplier and perform the multiplications sequentially; between these extremes the user can fine tune algorithm throughput versus resource usage.With a reuse factor of one, the computation is fully parallel. With a reuse factor of $R$, $\frac{1}{R}$ of the computation is done at a time with a factor of $\frac{1}{R}$ fewer multipliers.


The total number of multiplications required to infer a given LSTM layer using 16-bit is:
\begin{align*}
 DSP_{layer} &= \frac{4\times Lx \times Lh}{R_x} + \frac{4\times Lh^2 }{R_h} + 4\times Lh  \tag{3} \label{eq:3}  \\
 DSP_{model} &= \sum_{layer = 1}^{N} DSP_{layer} \leq DSP_{total}    \tag{4} \label{eq:4}
\end{align*}
Compared with the number of multipliers used in LSTM gates, the one required in the LSTM tail unit is small so the $R_t$ is set to 1. Otherwise, $\frac{4\times Lh}{R_t}$ should be used in~\eqnref{eq:3}. Besides, since the LSTM cell status, $c_{t-1}$, is represented in 32-bit, the $f_t \times c_{t-1}$ in the LSTM tail needs two Xilinx DSPs to implement one multiplier. Thus, the LSTM tail unit consumes $4\times Lh$ DSPs. The activation function sigmoid is implemented using BRAM-based lookup tables with a range of precomputed input values. The hyperbolic tangent function is implemented as piecewise linear function~\cite{moons2017minimum, azari2019energy} to reduce the latency. 
In the next subsection, we introduce our method for determining $R_x$ and $R_h$ with a given FPGAs. 

\subsection{Design space exploration}\label{subsec:explore}

FPGA multipliers are pipelined; therefore, the latency of one MVM computation, $LT_{mvm}$, is approximately
\begin{align*}
 LT_{mvm} &= LT_{mult} + (R-1)\times II_{mult}    \tag{5} \label{eq:5} 
\end{align*}
where $LT_{mult}$ is the latency of the multiplier, $II_{mult}$ is the initiation interval of the multiplier, which is one cycle in this work. \eqnref{eq:5} is approximate because, in some cases, additional cycles could be introduced for signal routing. Besides, the Vivado HLS tool will replace a multiplier by an adder when the corresponding weight is simple. 

As we discussed in~\secref{sec:design}, the optimal case is that the two sub-layers in an LSTM layer have the same II, which results in~\eqnref{eq:6}. 
\begin{align*}
II_{sublayer} &= LT_{mvm\_x} = LT_{mvm\_h} + LT_{\sigma} + LT_{tail}  \tag{6} \label{eq:6} 
\end{align*}
where $LT_{mvm\_x}$ and $LT_{mvm\_h}$ are the latencies of the MVM units involving input vectors $x$ and hidden vectors $h$ respectively. $LT_{\sigma}$ is the latency of the sigmoid function and $LT_{tail}$ is the latency of the LSTM tail unit. These units are shown in~\figref{fig:lstm_tran}. 
If we substitute the~\eqnref{eq:5} into~\eqnref{eq:6} and then we get
\begin{align*}
 R_x &= R_h + LT_{\sigma} + LT_{tail}.  \tag{7} \label{eq:7} 
\end{align*}

The architecture designed in this section serves as a baseline to deploy our methodology, whose goal is to find Pareto-optimal sets of reuse factors of the proposed accelerator to achieve a good trade-off between our design objectives, which are hardware resources, energy, and performance. To achieve low latency, the reuse factors should be as small as possible since when they decrease the parallelism increases, leading to high throughput. However, when reuse factors decrease, the required hardware resources increase and may easily exceed the number of total hardware resources on an FPGA. If we substitute the~\eqnref{eq:7} and~\eqnref{eq:3} into~\eqnref{eq:4}, we can get a quadratic inequality of $R_h$, which gives the minimum $R_h$ for a given number of DSPs.

\begin{figure}
\begin{center}
\includegraphics[width=0.9\linewidth]{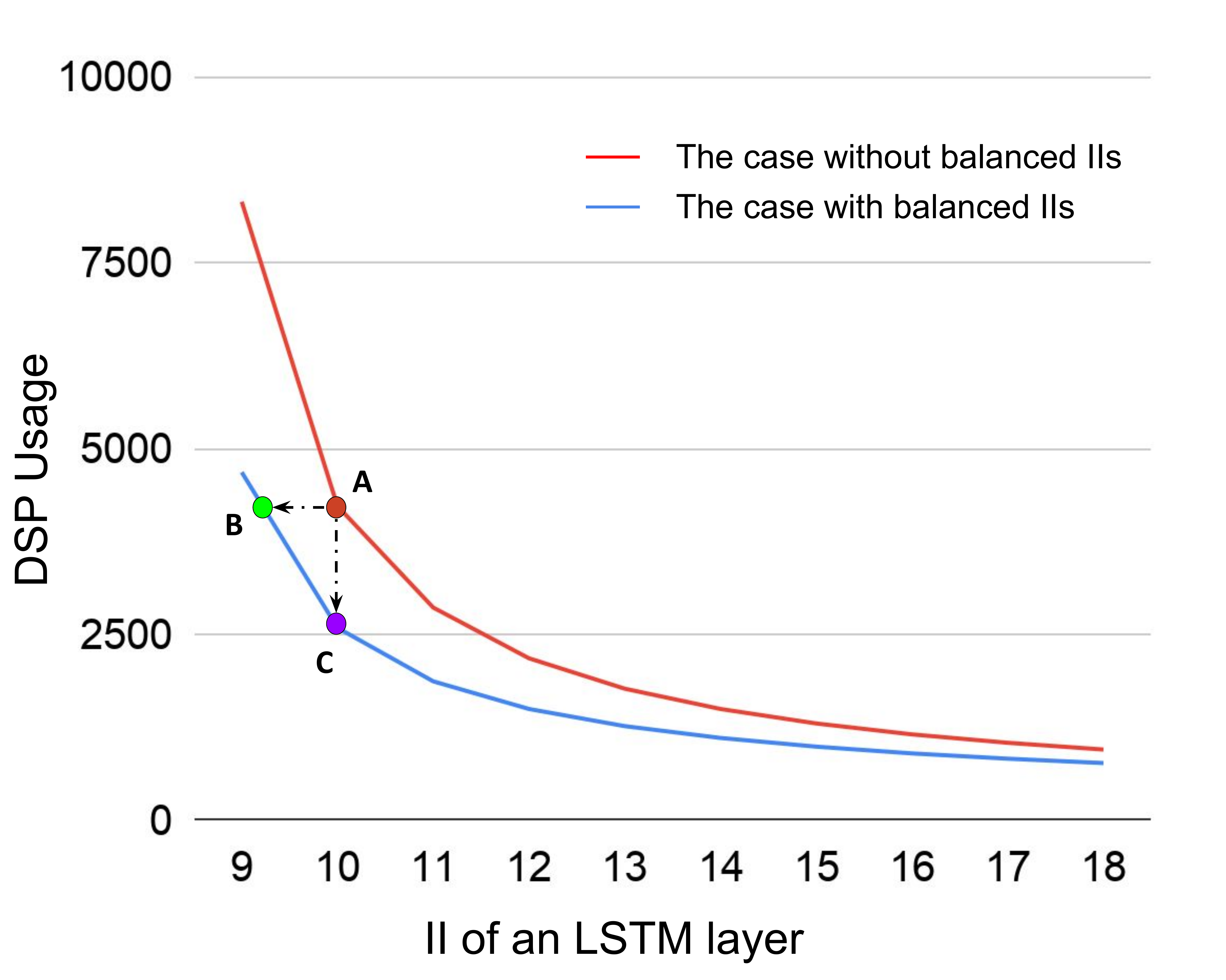}
\end{center}
   \caption{Pareto frontier}
\label{fig:pareto}
\end{figure}

\figref{fig:pareto} illustrates the exploration results of an LSTM layer with $(Lx, Lh) = (32,32)$ and different values of reuse factors, which are from 1 to 10. The red line represents the cases with the same $R_x$ and $R_h$. The blue line shows the cases with balanced IIs, where $R_x$ and $R_h$ meet the constraint in~\eqnref{eq:7}. For simplicity, $LT_{\sigma}$ is set to 3 and the $LT_{tail}$ is 5. Please note that $LT_{\sigma}$ and $LT_{tail}$ are both system dependent and can vary depending on clock frequency and FPGA devices. After balancing IIs, the Pareto frontier moves from red line to blue line. With the proposed technique, we can achieve a same II with less DSP usage (from point A to point C) or we can achieve a better II (from point A to point B) as shown in~\figref{fig:pareto}.

\section{Evaluation and Analysis}
This section presents the performance of the RNN models developed for gravitational wave detection on two generations of Xilinx FPGAs demonstrating the scalability of the proposed optimization.

\subsection{Experimental setup}
Simulated gravitational waves are generated using the GGWD library \cite{GGWD}. Noise is generated at a specified power spectral density (PSD) to mimic normal detector background conditions using PyCBC ~\cite{alex_nitz_2020_3993665}. This approach to simulated data generation ignores glitches, blips, and other transient sources of detector noise, though this algorithm can be re-purposed for identifying these detector glitches with unsupervised methods. Signal events are generated simulating GW production from compact binary coalescences using PyCBC~\cite{alex_nitz_2020_3993665}, which itself uses algorithms from LIGO's LAL Suite~\cite{lalsuite}. Signal events containing GWs were created overlaying simulated GWs, with the SEOBNRv4 Approximant, on top of detector noise. This provides an analogous situation to a real GW, in which the strain from the incoming wave is recorded in combination with the normal detector noise. Data are then whitened and band-passed, then normalized. 
The training set has 240K gravitational wave events. 
The validation set and test set have 60k and 50k events respectively. To study the performance and limitations of the proposed optimizations and hardware architecture, the designs are implemented using Vivado HLS 19.2. Two generations of Xilinx FPGAs, the ZYNQ 7045 and U250, are evaluated and compared with previous work. 




\begin{figure}
\begin{center}
\includegraphics[width=0.9\linewidth]{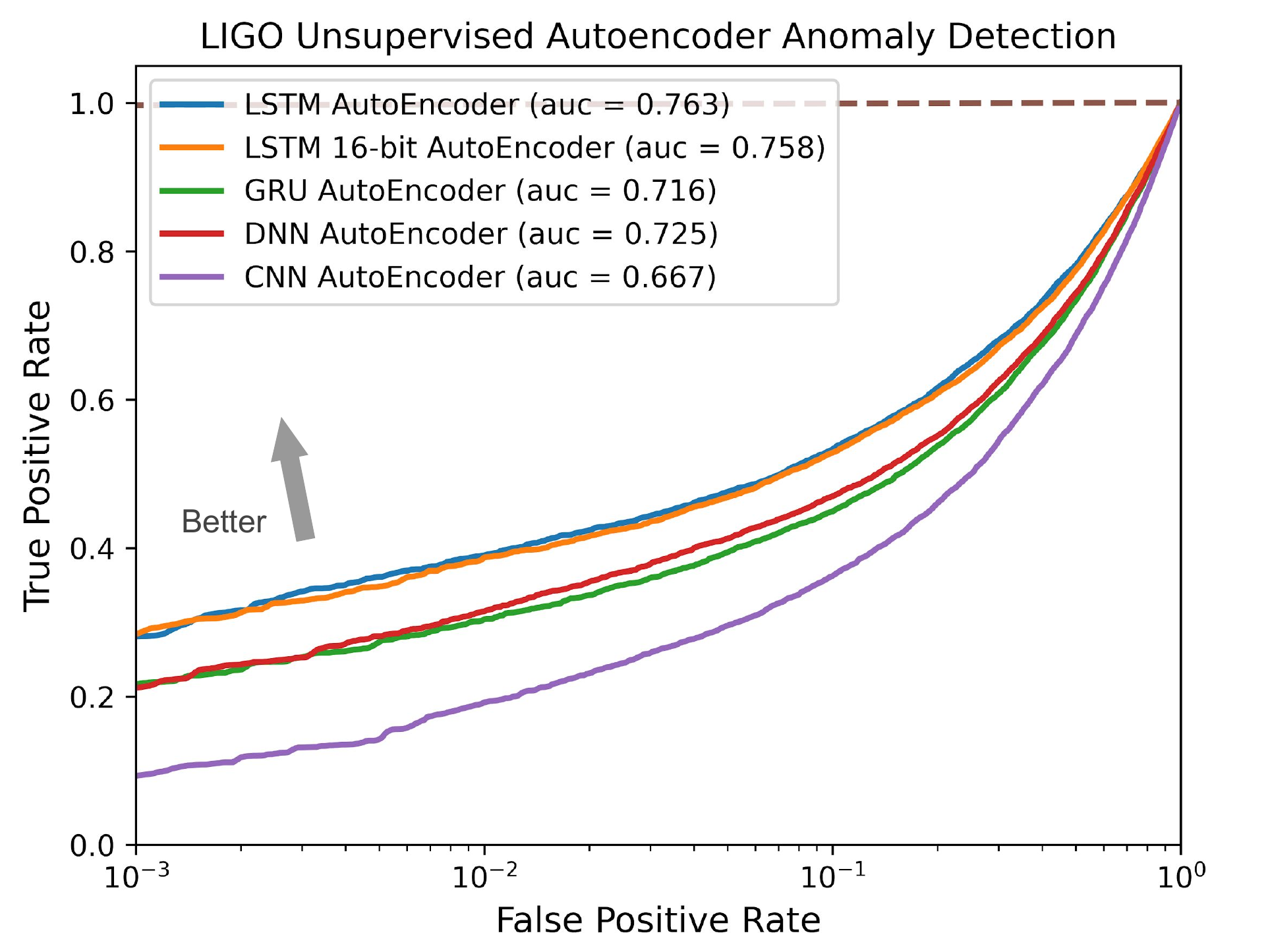}
\end{center}
   \caption{AUCs and ROC curves for various autoencoders}
\label{fig:auc}
\end{figure}

\subsection{Model accuracy}

To quantify the performance of the autoencoders for anomaly detection implemented by various neural networks, we use the AUC metric, or area under the Receiver Operating Characteristic (ROC) curve, as shown in~\figref{fig:auc}, with higher AUC corresponding to better performance. The default timestep~\cite{ligo_anomaly} of 100 is used. AUC is a common metric for evaluating models as it is classification-threshold-invariant. The threshold for flagging an anomaly by its loss spike can be calculated by setting a false positive rate (FPR) on noise events. The higher the threshold for detecting an anomaly, the lower the FPR will be. This threshold can be used to calculate the corresponding true positive rate (TPR) on signal events. 
We observe that the LSTM-based autoencoder has the highest AUC, and hence the best performance, among the unsupervised designs~\cite{ligo_anomaly} with various NN layers, including GRU, CNN and DNN. Additionally, Qkeras~\cite{coelho2006automatic} is used to quantize the LSTM-based autoencoder to 16-bit.
We find this precision to have a negligible effect on the NN performance.


\subsection{Performance and efficiency comparison}


To illustrate the benefits of our proposed approach, two LSTM-based autoencoders are evaluated. The first one is a small autoencoder which has the same architecture as the one used in gravitational wave detection described in~\secref{sec:autoencoder} but only has two LSTM layers, each having 9 hidden units. The results are shown in~\tabref{table:cmp_fpga}. It is running at 100MHz with 8 timesteps. The weights and input are 16~bits.
The bias and LSTM cell status are both 32~bits to keep the accuracy. 
To achieve the lowest latency, the reuse factors should be set to one so that all the operations are unrolled, e.g., the design Z1 in~\tabref{table:cmp_fpga}. However the required number of DSPs exceed the one of the total DSPs on this FPGA. One may increase the re-use factor from one to two to fit the design into this FPGA device. However the cost is that now the timestep loop initiation interval, $ii_{layer}$, increases by one cycle which results in $TS$ cycles increase for the layer II, e.g., the design Z2 in~\tabref{table:cmp_fpga}. 
However, it is not necessary to fully unroll all units in order to achieve the lowest latency. Some hardware resources can be saved from the units which do not require full unrolling and can be allocated to the other units which are dominating to achieve low latency. 
\begin{table}[t]
\centering
\caption{Performance comparison of the FPGA designs}
\label{table:cmp_fpga}
\renewcommand{\arraystretch}{1.1}
\begin{threeparttable}
\centering
\begin{tabular}{|c|c|c|c|c|c|c|}
\hline
 & Z1 & Z2 & Z3 & U1 & U2 & U3 \\ \hline
FPGA & \multicolumn{3}{c|}{ Zynq 7045} & \multicolumn{3}{c|}{ U250} \\ \hline

\begin{tabular}[c]{@{}c@{}}DSP \\ total \end{tabular}
& \multicolumn{3}{c|}{ 900 } & \multicolumn{3}{c|}{ 12,288 } \\ \hline
$R_h$  & 1 & 2 & 1 & 1 & 1 & 4  \\ \hline
$R_x$  & 1 & 2 & 9 & 1 & 9 & 12\\ \hline
\begin{tabular}[c]{@{}c@{}}LUT  \\ used\end{tabular} & 
\begin{tabular}[c]{@{}c@{}}45k \\ (21\%) \end{tabular} & 
\begin{tabular}[c]{@{}c@{}}45k \\ (21\%) \end{tabular} &
\begin{tabular}[c]{@{}c@{}}43k \\ (20\%) \end{tabular} &
\begin{tabular}[c]{@{}c@{}}449k \\ (26\%) \end{tabular} &
\begin{tabular}[c]{@{}c@{}}463k \\ (27\%) \end{tabular} & 
\begin{tabular}[c]{@{}c@{}}516k \\ (30\%) \end{tabular} \\ \hline

\begin{tabular}[c]{@{}c@{}}DSP  \\ used \end{tabular} & 
\begin{tabular}[c]{@{}c@{}}1,058 \\ (118\%) \end{tabular} & 
\begin{tabular}[c]{@{}c@{}}578 \\ (64\%)\end{tabular} &
\begin{tabular}[c]{@{}c@{}}744 \\ (83\%) \end{tabular} &
\begin{tabular}[c]{@{}c@{}}11,123 \\ (91\%) \end{tabular} &
\begin{tabular}[c]{@{}c@{}}9,021 \\ (73\%) \end{tabular} &
\begin{tabular}[c]{@{}c@{}}2,713 \\ (22\%) \end{tabular} \\ \hline

\begin{tabular}[c]{@{}c@{}}$ii_{layer} $  \\ cycles \end{tabular}  & 
9 & 10 & 9 & 12 & 12 & 13\\ \hline
\begin{tabular}[c]{@{}c@{}}$II_{layer} $  \\ cycles \end{tabular}  & 
72 & 80 & 72 & 96 & 96 & 104  \\ \hline
\end{tabular}
\vspace{0.2cm}
\end{threeparttable}
\end{table}

\begin{figure}
\begin{center}
\includegraphics[width=0.9\linewidth]{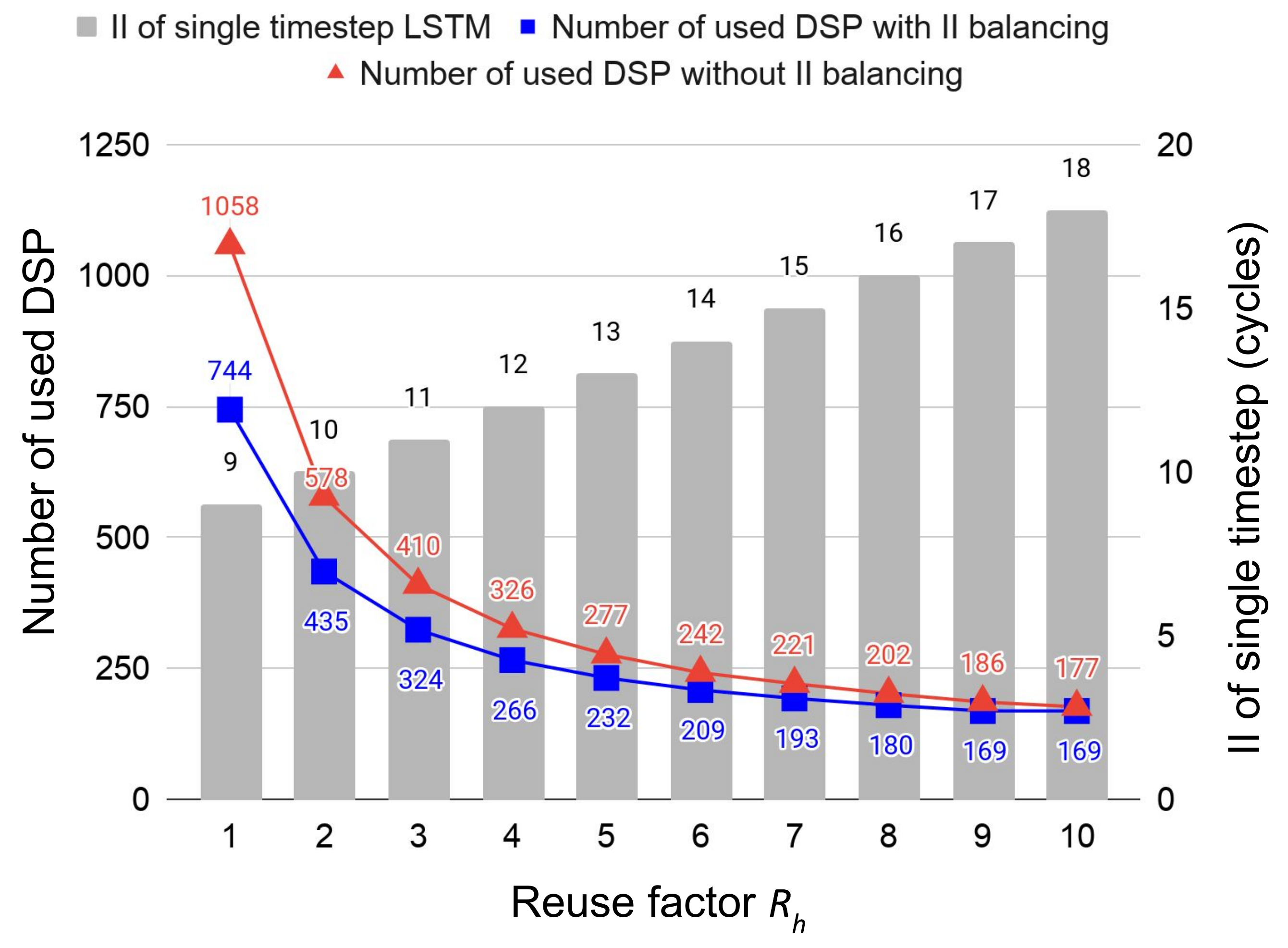}
\end{center}
   \caption{Initiation intervals and DSP numbers using various reuse factor $R_h$ on Zynq 7045}
\label{fig:result_z}
\end{figure}

With the proposed balancing of IIs, some of the DSPs resources can be rearranged from implementing $mvm\_x$ to $mvm\_h$ to achieve lower latency, e.g., the design Z3. So this design can still achieve the lowest II like the case with full unrolling, and it is still able to fit in this FPGA device as shown in~\tabref{table:cmp_fpga}, showing the benefits of balanced IIs. 
Besides, with heterogeneous reuse factors, the parallelism of the design can be fine-tuned to make the trade-off between latency, throughput and FPGA hardware resources as shown in~\figref{fig:result_z}. With the balanced II, the number of DSPs can be reduced up to 42\% while achieving the same IIs.

Besides, to show the adaptability of our technique, the nominal autoencoder~\cite{ligo_anomaly} developed for gravitational wave detection is implemented using a larger FPGA, U250, running at 300MHz with 8 timesteps. It has four LSTM layers which have a number of hidden units equal to 32, 8, 8, 32 respectively and one TimeDistributed dense layer before the output. 
Since the U250 has 12,288 DSPs, the whole fully unrolled autoencoder can be fit into this FPGA with both $R_x$ and $R_h$ set to one, shown as the design U1 in~\tabref{table:cmp_fpga}. With our technique of balancing IIs, the DSPs of the design U2 can be reduced by 2102 while achieving the same design IIs and same design throughput. After HLS synthesis, the II is slightly larger than the one estimated by the performance model since the DSP usage is very high and some additional cycles are incurred for signal routing. The design U3 is an interesting version with reuse factors ($R_h$, $R_x$) as (4, 12). It achieves a slightly worse II, as shown in~\tabref{table:cmp_fpga}, however it consumes 3.3 and 4.1 times less DSPs than design U2 and design U1 respectively. 
Sometimes, the user may only care about the latency of the LSTM running on the FPGAs, then they can just take the point that gives them the lowest latency with most resources. 
However, if the user can bear with a slightly reduced latency then they can choose a smaller and cheaper FPGA as shown in~\tabref{table:cmp_fpga}. One can choose between using less resources but increasing latency and vice versa.
Please note because of the data dependence, the $ii_{layer}$ could be hard to optimize to 1. However, it could be further optimized to a smaller value using fast multipliers or fast activation functions. We leave that for future work since it has a limited impact on the conclusions we draw from our study in this paper.

To compare the performance of the proposed design on FPGA with other platforms, we implement the same LSTM-based autoencoder on Intel CPU and NVIDIA GPU. The AVX2 vector instructions are enabled for the CPU while the CuDNN libraries are enabled for the GPU. 
Compared with the designs running on CPU and GPU, our FPGA design 
runs much faster, as shown in~\tabref{table:cmp_cpugpu}. 
We are processing each inference sequentially (batch 1) since requests need to be processed as soon as they arrive. The GPUs provide large throughput by running many parallel inferences but may not perform well when the batch is small, especially there are data dependencies in LSTMs. However, FPGAs work fast on a single inference with a fully unrolled tailor-made design.

\begin{table}[t]
\centering

\caption{Latency comparison of the FPGA design versus CPU and GPU}
\label{table:cmp_cpugpu}
\renewcommand{\arraystretch}{1.2}
\begin{threeparttable}
\centering
\begin{tabular}{|c|c|c|c|}
\hline
 & CPU & GPU & This work \\ \hline
Platform 
& \begin{tabular}[c]{@{}c@{}}Intel E2620 \end{tabular} 
& \begin{tabular}[c]{@{}c@{}}TITAN X \end{tabular} 
& \begin{tabular}[c]{@{}c@{}}U250 \end{tabular} 
\\ \hline
Precision & F32 & F32 & 16 Fixed \\ \hline
Latency & 39.7 ms & 32.1 ms & 0.40 us  \\ \hline
\end{tabular}
\vspace{0.1cm}
\end{threeparttable}
\end{table}


Some other HLS-based RNN/LSTM accelerators on FPGAs are compared with ours in~\tabref{table:cmp_perf}. In this table, we focus on latency since the throughput, power or power efficiency of the other designs are not reported.
Our design achieves 4.92 to 12.4 times lower latency compared to the state-of-the-art FPGA designs targeting anomaly detection. 
Our single-layer design, with a similar amount of DSP resources to another design~\cite{rao2020implementation}, is 3.9 times faster as shown in~~\tabref{table:cmp_perf}. 
Note that because of the structure of an autoencoder, the processing of the encoder and the decoder cannot be overlapped, which increases the end-to-end latency of the design. 
Nevertheless, we still achieve better latency than the others which contain only one LSTM layer. 
Moreover, while the other designs report Vivado HLS synthesis latency, we report the RTL co-simulation latency which is likely to be more accurate.

\begin{table}[t]
\centering

\begin{threeparttable}
\caption{Comparison with previous FPGA-based LSTM designs for anomaly detection and physics }
\vspace{0.2 cm}
\label{table:cmp_perf}
\renewcommand{\arraystretch}{1.1}
\centering

\begin{tabular}{c|c|c|c|c}
\hline
& \cite{lee2018long}, 2018
& \cite{rao2020implementation}, 2020
& This work
& This work
\\ \hline
FPGA 
& \begin{tabular}[c]{@{}c@{}}Kintex7\\ K410T\end{tabular}
& KU115
& U250 
& U250 
\\ \hline
Model 
& \begin{tabular}[c]{@{}c@{}}Single \\ Layer\end{tabular}
& \begin{tabular}[c]{@{}c@{}}Single\\ Layer\end{tabular} 
& \begin{tabular}[c]{@{}c@{}}Single\\ Layers\end{tabular} 
& \begin{tabular}[c]{@{}c@{}}Four\\ Layers\end{tabular} 
 \\ \hline
\begin{tabular}[c]{@{}c@{}}Application \\ Domain\end{tabular}
& \begin{tabular}[c]{@{}c@{}}Anomaly \\ Detection\end{tabular}
& \begin{tabular}[c]{@{}c@{}}Physics \end{tabular}
& \begin{tabular}[c]{@{}c@{}} - \end{tabular}
& \begin{tabular}[c]{@{}c@{}}Anomaly \\ Detection\end{tabular}
\\ \hline
\begin{tabular}[c]{@{}c@{}}LSTM hidden \\ units $Lh$\end{tabular}
 & 32 & 16  & 32 & 32,8,8,32  
\\ \hline
DSPs
 & 1091  & 2374 & 2221 & 9021  
\\ \hline
Preci. (bits) 
 & 16 fixed &  16 fixed &  16 fixed &  16 fixed  \\ \hline

Freq. (MHz)
 & 155  & 200  & 300 & 300 \\ \hline

Latency (us) 
 & 4.27 & 1.35  &  0.343 &  0.867  \\ \hline


\end{tabular}
\vspace{ 0.2 cm}
\end{threeparttable}
\end{table}

\section{Related Work} 
A latency-optimized LSTM-based anomaly detection is proposed in~\cite{lee2018long} on FPGAs and we achieve 4.9 times faster than it. \cite{duarte2018fast} proposes the HLS4ML tool and introduces a deep FC-layer model for substructure-based jet tagging in LHC physics. \cite{rao2020implementation} introduces HLS LSTMs for the same physics problem. 

Partitioning FPGA resources to improve throughput has been studied for CNNs ~\cite{zhang2018dnnbuilder, nakahara2020high, shen2017maximizing, zhang2020dnnexplorer}, but they do not touch the RNNs and the recurrent nature and data dependency in RNN computations which are absent in CNNs. The FiC-RNN~\cite{sun2020fic} proposes to accelerate multi-layer RNNs using an FPGA cluster, in which each RNN layer occupies a single FPGAs. The authors in~\cite{peng2019exploiting} put each LSTM layer on each multi-core to achieve coarse grained pipelining. In~\cite{rybalkin2018finn, que2019efficient, boutros2020beyond,que2020mapping}, the batching technique is used to improve the hardware throughput and utilization for LSTM inferences. However, latency can suffer since different inputs may not come at the same time, meaning that a newly arrived request has to wait until the batch is formed, which imposes a significant latency penalty. Some of the previous studies~\cite{han2017ese, chen2018clink, cao2019efficient, shi2019lstm, nan2020dc,chen2020blink} are focusing on weight pruning and model compression to achieve good performance and efficiency. Some researchers use low bitwidth, even binarized, datapaths \cite{rybalkin2018finn, nurvitadhi2016BNN, rybalkin2020massive} and investigate the trade-off between precision and performance. These studies are orthogonal to our proposed approach and hardware architecture. These techniques can be complementary to our approach to achieve even lower latency of RNN inferences on FPGAs.




\section{Conclusions and Future Work}

This paper aims to pioneer new data analysis architectures to support next-generation low-latency anomaly detection on time series data, relevant to many fundamental physics experiments including gravitational wave detection. We present a novel approach for minimizing the initiation intervals for the execution of a multi-layer LSTM network by optimizing the reuse factors for each layer. Results show latency reduction of up to 12.4 times over the existing FPGA-based LSTM design. Current and future work includes exploring the use of new FPGA resources such as the AI Engines~\cite{xilinx_white} and the AI Tensor Blocks~\cite{langhammer2021stratix}, and incorporating the proposed approach into the design of the data analysis architecture for next-generation gravitational wave detectors.

\section*{Acknowledgement}
The support of the United Kingdom EPSRC (grant numbers
EP/L016796/1, EP/N031768/1, EP/P010040/1, and EP/S030069/1), CERN and Xilinx is gratefully acknowledged. We thank Prof. Zhiru Zhang and Yixiao Du for their help and advice. 

\scriptsize
\bibliographystyle{IEEEtran}
\bibliography{main-bibliography}

\end{document}